\documentclass[sigconf]{acmart}

\AtBeginDocument{%
  }

\copyrightyear{2024}
\acmYear{2024}
\setcopyright{acmlicensed}
\acmConference[MM '24]{Proceedings of the 32nd ACM International Conference on Multimedia}{October 28-November 1, 2024}{Melbourne, VIC, Australia}
\acmBooktitle{Proceedings of the 32nd ACM International Conference on Multimedia (MM '24), October 28-November 1, 2024, Melbourne, VIC, Australia}
\acmDOI{10.1145/3664647.3680676}
\acmISBN{979-8-4007-0686-8/24/10}

\usepackage{algorithm}
\usepackage{balance}
\usepackage{algorithmic}
\usepackage{multirow}
\usepackage{float}               
\usepackage{subfig} 
\usepackage{bbding}

\begin{document}

\title{ZePo: Zero-Shot Portrait Stylization with Faster Sampling}

\author{Jin Liu}
\orcid{0009-0007-7320-5479}
\affiliation{%
  \institution{SIST, ShanghaiTech University}
  \institution{NLPR \& MAIS, Institute of Automation, CAS}
  \city{Shanghai \& Beijing}
  \country{China}
}
\email{liujin2@shanghaitech.edu.cn}

\author{Huaibo Huang}
\authornote{Corresponding author.}
\orcid{0000-0001-5866-2283}
\affiliation{%
  \institution{NLPR \& MAIS, Institute of Automation, CAS}
  \city{Beijing}
  \country{China}
}
\email{huaibo.huang@cripac.ia.ac.cn}

\author{Jie Cao}
\orcid{0000-0001-6368-4495}
\affiliation{
\institution{NLPR \& MAIS, Institute of Automation, CAS}
\city{Beijing}
\country{China}
}
\email{jie.cao@cripac.ia.ac.cn} 

\author{Ran He}
\orcid{0000-0002-3807-991X}
\affiliation{
\institution{NLPR \& MAIS, Institute of Automation, CAS}
\city{Beijing}
\country{China}
}
\email{rhe@nlpr.ia.ac.cn}
\renewcommand{\shortauthors}{Jin Liu, Huaibo Huang et al.}

\begin{abstract}
Diffusion-based text-to-image generation models have significantly advanced the field of art content synthesis. However, current portrait stylization methods generally require either model fine-tuning based on examples or the employment of DDIM Inversion to revert images to noise space, both of which substantially decelerate the image generation process. To overcome these limitations, this paper presents an inversion-free portrait stylization framework based on diffusion models that accomplishes content and style feature fusion in merely four sampling steps. We observed that Latent Consistency Models employing consistency distillation can effectively extract representative Consistency Features from noisy images. To blend the Consistency Features extracted from both content and style images, we introduce a Style Enhancement Attention Control technique that meticulously merges content and style features within the attention space of the target image. Moreover, we propose a feature merging strategy to amalgamate redundant features in Consistency Features, thereby reducing the computational load of attention control. Extensive experiments have validated the effectiveness of our proposed framework in enhancing stylization efficiency and fidelity.
The code is available at 
\url{https://github.com/liujin112/ZePo}.
\end{abstract}

\ccsdesc[500]{Computing methodologies~Image manipulation}
\keywords{Portrait Stylization, Diffusion Model, Zero-Shot}

\begin{teaserfigure}
  \includegraphics[width=\textwidth]{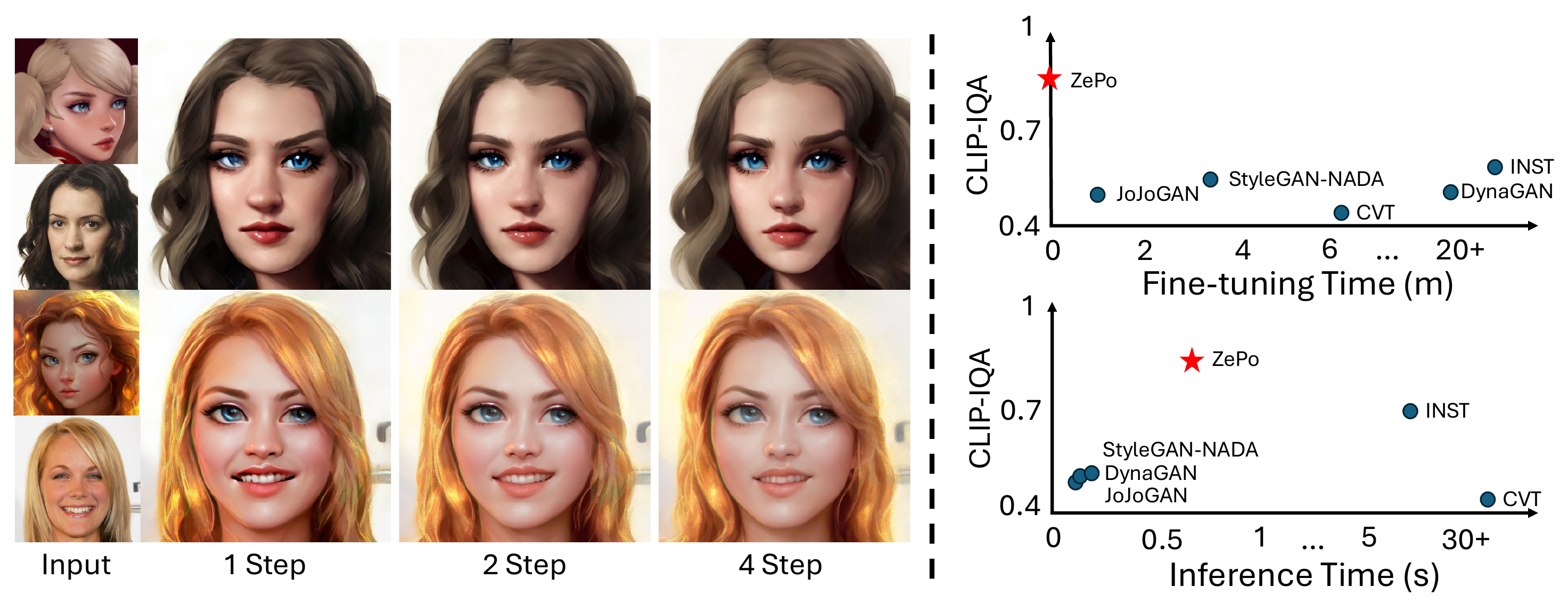}
  \caption{The proposed zero-shot portrait stylization framework \textbf{ZePo} can directly synthesize stylized facial images with very few sampling steps (including 1, 2, and 4 steps) (left), where the images synthesized in 4 steps have the best overall quality as measured by the CLIP-IQA metric. Moreover, our method does not require model fine-tuning, and with 4-step sampling, the inference time is only about 0.6 seconds (right).}
  \label{fig:teaser}
\end{teaserfigure}
\maketitle
\section{Introduction}
Portrait stylization involves the transfer of an art style from a reference portrait to a standard facial photograph. Traditional methods for portrait stylization \cite{chongJojoganOneShot2022,songAgileganStylizingPortraits2021,liStyOStylizeYour2023,zhangInversionbasedStyleTransfer2023,chengGeneralImagetoImageTranslation2023} typically involve fine-tuning a pre-trained generative model \cite{goodfellowGenerativeAdversarialNetworks2020,rombachHighresolutionImageSynthesis2022}, such as StyleGAN \cite{karrasStylebasedGeneratorArchitecture2019} or Stable Diffusion \cite{rombachHighresolutionImageSynthesis2022}, using various reference art portraits. However, these approaches necessitate considerable time for model fine-tuning and additional storage space to accommodate the models that have been fine-tuned for each distinct style image.

To overcome the limitations mentioned above, recent studies have investigated a tuning-free method \cite{liuPortraitDiffusionTrainingfree2023,dengZeroshotStyleTransfer2023,chungStyleInjectionDiffusion2024} that leverages self-attention features from both content and reference images during the generation process through Attention Control \cite{caoMasaCtrlTuningFreeMutual2023}, enabling zero-shot portrait stylization. Despite this advancement, the method struggles with slow image generation speeds. On the one hand, the diffusion model requires an extensive sampling process to iteratively denoise Gaussian noise. On the other hand, to accurately reconstruct the content and reference images, this method often depends on the protracted DDIM Inversion \cite{songDenoisingDiffusionImplicit2020} process, which is necessary to obtain a sequence of intermediate anchors for image reconstruction. Additionally, the manually customized Attention Control \cite{caoMasaCtrlTuningFreeMutual2023} operation exacerbates the situation by involving excessive computations of the redundant self-attention mechanism, further impeding the image generation speed.

In this work, we introduce \textbf{ZePo}, a \textbf{Ze}ro-shot \textbf{Po}rtrait Stylization framework, to address the aforementioned challenges. Regarding the issue of slow sampling speeds, one intuitive solution is to employ high-order numerical ODE solvers \cite{baoAnalyticDPMAnalyticEstimate2021,luDPMSolverFastSolver2022,zhaoUnipcUnifiedPredictorcorrector2024} to decrease the number of sampling steps required for image generation.
However, these methods, which leverage high-order ODE approximations, necessitate multiple network function evaluations (NFEs) and achieve only a marginal reduction in actual sampling time. Moreover, these ODE solvers do not integrate well with the intermediate anchors established by DDIM Inversion, which restricts their applicability for this particular method. Therefore, rather than relying on high-order ODE samplers, we propose the use of accelerated distillation of pre-trained models, specifically Latent Consistency Models (LCMs) \cite{luoLatentConsistencyModels2023}, to expedite the image synthesis process. Additionally, to obviate the need for DDIM Inversion, our findings indicate that LCMs can directly extract representative consistency features from noised images. Building on this capability, we suggest a method to directly extract consistency features from noisy reference and content images. These features are then seamlessly incorporated during the generation process of the target image, resulting in a more efficient and streamlined stylization approach. 

To address the issue of speed reduction due to redundant computations in conventional Attention Control methods, we introduce the Style Enhancement Attention Control (SEAC). SEAC begins by integrating the redundant consistency features from both the source and reference images. Subsequently, it concatenates these merged features and maps them as key and value features within the self-attention space. To modulate the degree of image stylization, the key features of the reference image are multiplied by a Style Enhancement coefficient. Consequently, the attention map, calculated using the query features from the target image and the merged key features, can adaptively select the value features from both the content and reference images. This method not only increases the computational speed of Attention Control but also mitigates the issue of query confusion, enhancing the precision and efficiency of the stylization process.

Ultimately, as illustrated in Figure \ref{fig:teaser} (left), our method demonstrates the capability to synthesize stylized portraits using no more than four sampling steps, significantly enhancing both the speed and practicality of portrait stylization using diffusion models.
Through extensive experimentation, we have demonstrated the advantages of our ZePo framework in rapid stylized portrait synthesis. As illustrated in Figure \ref{fig:teaser} (right), ZePo does not require additional fine-tuning time, and it achieves the optimal CLIP-IQA score while reducing the inference time to just 0.6 seconds using a 4-step sampling process.

To summarize, we make the following key contributions:

(i) We introduce ZePo, a new inversion-free portrait stylization framework that requires as few as one sampling step to synthesize high-quality stylized portraits.

(ii) We propose a novel attention control mechanism, termed Style Enhancement Attention Control, which leverages redundant feature fusion to enhance the speed of self-attention computations and can adaptively select value features from source and reference images.

(iii) We demonstrate from both quantitative and qualitative perspectives that our method surpasses existing state-of-the-art baselines, achieving a significantly better balance between preserving source content information and enhancing image stylization.
\section{Related works}

\subsection{Few-Shot Face Stylization}

Early methods of face stylization \cite{isolaImagetoImageTranslationConditional2017,wangHighresolutionImageSynthesis2018,zhuMultimodalImagetoImageTranslation2017,zhuUnpairedImagetoimageTranslation2017,liuUnsupervisedImagetoimageTranslation2017,kimUGATITUnsupervisedGenerative2019,li2020disentangled,wang2021attentional,xie2022artistic,cui2024instastyle} required extensive data sampling for training image-to-image translation models, resulting in substantial training costs. 
To reduce training costs and leverage pre-trained models, few-shot face stylization emerged. This technique involves fine-tuning a pre-trained StyleGAN model \cite{karrasStylebasedGeneratorArchitecture2019,karrasAnalyzingImprovingImage2020,karrasTrainingGenerativeAdversarial2020} with limited target images, known as GAN-adaptation \cite{wangTransferringGansGenerating2018,noguchiImageGenerationSmall2019,wangMineganEffectiveKnowledge2020,robbFewshotAdaptationGenerative2020,zhaoLeveragingPretrainedGANs2020,yangOneShotGenerativeDomain2023,ZhouDeformableOne2024}.
Toonify \cite{pinkneyResolutionDependentGan2020} pioneered this by fine-tuning a StyleGAN model with a few cartoon samples and interpolating the fine-tuned model's weights with the original model's to create cartoon-styled faces. \cite{liFewshotImageGeneration2020,ojhaFewshotImageGeneration2021} added regularization terms to the latent space to prevent overfitting during fine-tuning with few samples. 
methods like AgileGAN \cite{songAgileganStylizingPortraits2021} and DualStyleGAN \cite{yangPasticheMasterExemplarBased2022} introduced frameworks and paths to enhance consistency and efficiency, but still needed many images. 
JoJoGAN \cite{chongJojoganOneShot2022} advanced this with a one-shot face stylization method using a reference image to generate a style-mixed paired dataset, enhancing utility in limited-sample environments. \cite{zhangGeneralizedOneshotDomain2022} proposed a novel one-shot adaptation method for face stylization, separating style transformation from identity transformation for more natural outcomes. StyleDomain \cite{alanovStyleDomainEfficientLightweight2023} introduced a parameter-efficient method modifying style vectors in the Style Space to adapt pre-trained models to new domains with minimal resources. \cite{ZhouDeformableOne2024} used a single real-style paired reference for style direction in the DINO-ViT \cite{caronEmergingPropertiesSelfsupervised2021} feature space for precise fine-tuning. With the rapid development of multimodal learning \cite{he2014robust,he2015cross}, CLIP-based methods \cite{patashnikStyleclipTextdrivenManipulation2021,kwonClipstylerImageStyle2022,cheferImagebasedClipguidedEssence2022,galStyleGANNADACLIPguidedDomain2022} have explored zero-shot GAN adaptation for image stylization based on textual or image prompts through CLIP \cite{radfordLearningTransferableVisual2021}, achieving strong generalization.

\subsection{Diffusion-Based Style Transfer}

\begin{figure*}[!h]
        \centering
        \includegraphics[width=\linewidth]{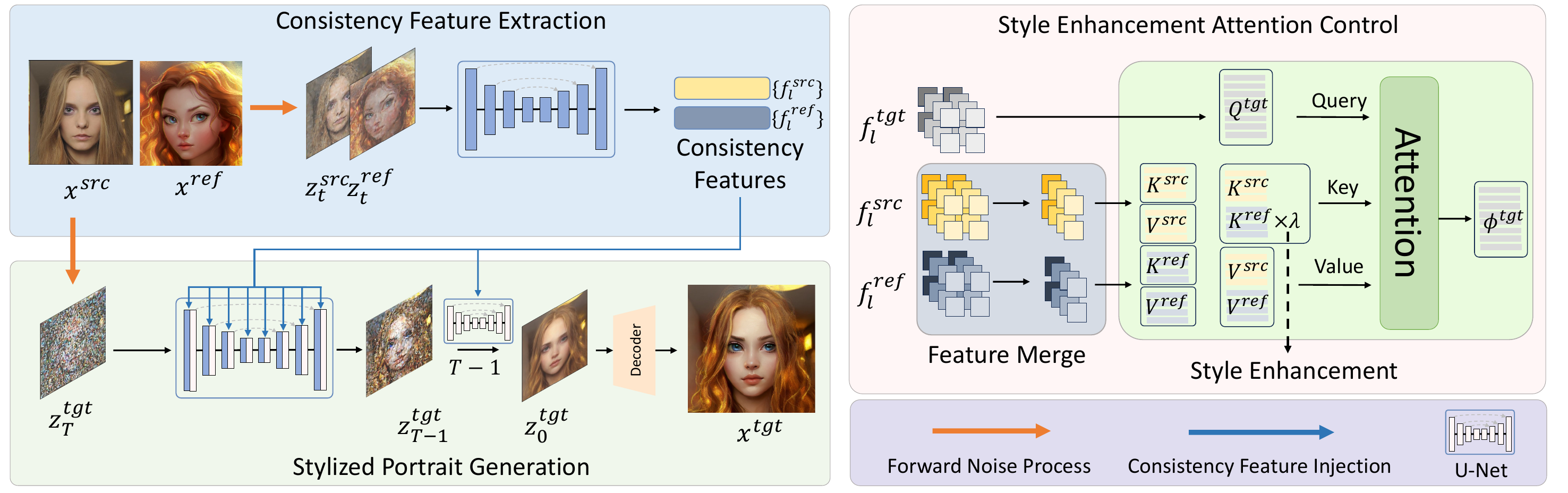}
        \caption{The overall framework of \textbf{ZePo}. The framework is divided into two stages. The first stage involves the extraction of consistency features, where multi-scale consistent features are extracted from the reference and source images with slight noise added. The second stage is the stylized image synthesis phase, where the source image, added with a moderate level of noise, is used as the input. In this phase, the Style Enhancement Attention Control module within the U-Net fuses the consistency features from both the reference and source images to synthesize a stylized portrait.} 
        \label{fig:framework}
\end{figure*}

Diffusion models \cite{sohl-dicksteinDeepUnsupervisedLearning2015,hoDenoisingDiffusionProbabilistic2020,songScoreBasedGenerativeModeling2020,songDenoisingDiffusionImplicit2020,dhariwalDiffusionModelsBeat2021} have gained prominence in generative models, especially with pre-trained text-to-image (T2I) models \cite{openaiDALL2023,rombachHighresolutionImageSynthesis2022,sahariaPhotorealisticTexttoimageDiffusion2022}. These models have popularized AI-generated art and spurred research into diffusion-based style transfer methods. Some methods utilize classifier guidance \cite{dhariwalDiffusionModelsBeat2021} and energy functions for guiding stylized image generation \cite{yuFreedomTrainingfreeEnergyguided2023,nairSteeredDiffusionGeneralized2023}. Techniques like \cite{kwonDiffusionbasedImageTranslation2022} use style loss from pre-trained models (e.g., DINO-ViT) to generate stylized images, while \cite{YangZeroShotContrastiveLoss2023} leverages CLIP-generated style loss. 
Fine-tuning approaches like LoRA \cite{huLoRALowRankAdaptation2021}, Textual Inversion \cite{galImageWorthOne2022}, and Dreambooth \cite{ruizDreamboothFineTuning2023} personalize diffusion models for specific styles. DDIM Inversion \cite{songDenoisingDiffusionImplicit2020,preechakulDiffusionAutoencodersMeaningful2022} helps generate stylized images by denoising in the noise space \cite{kimDiffusionCLIPTextGuidedDiffusion2022,zhangInversionbasedStyleTransfer2023, chengGeneralImagetoImageTranslation2023}, though it can slow down synthesis. Methods like \cite{choOneShotStructureAwareStylized2024} fine-tune Diffusion Autoencoders with optimized latent codes for precise control of content and style. ControlNet \cite{zhangAddingConditionalControl2023} and T2I-Adapter \cite{mouT2iadapterLearningAdapters2024} train style adapters for pre-trained T2I models, providing tailored style management.
Recent zero-shot stylization methods \cite{liuPortraitDiffusionTrainingfree2023,dengZeroshotStyleTransfer2023,chungStyleInjectionDiffusion2024} use pre-trained T2I diffusion models with attention control modules \cite{caoMasaCtrlTuningFreeMutual2023} to integrate content and style features but rely on DDIM Inversion, which slows down image synthesis.

\section{Preliminaries}

\subsection{Latent Diffusion Models}

Latent Diffusion Models (LDMs) \cite{rombachHighresolutionImageSynthesis2022} employ a diffusion model within the latent space of a pre-trained Variational Autoencoder (VAE) \cite{esserTamingTransformersHighresolution2021}. The encoder \(\mathcal{E}\) encodes images into latent codes \(z_0 = \mathcal{E}(x)\), while the decoder \(\mathcal{D}\) reconstructs images \(x = \mathcal{D}(z_0)\) from these codes.

The forward process of diffusion models operates as a Markov chain, incrementally introducing noise into the initial latent code \(z_0\). Due to the additive nature of Gaussian noise, this process is generally modeled as a single-step addition of noise, directly yielding the noisy latent code \(z_t\) at any given step \(t\):
\begin{gather}
    \label{eq:forward_ddpm}
    \mathbf{z}_{\mathbf{t}} = \sqrt{\alpha_{\mathbf{t}}}\mathbf{z}_0 + \sqrt{1-\alpha_{\mathbf{t}}}\boldsymbol{\epsilon}, \quad \boldsymbol{\epsilon} \sim \mathcal{N}(\mathbf{0}, \mathbf{I}),
\end{gather}
where \(\alpha_t\) is a predefined diffusion schedule. The reverse process of diffusion models constitutes an approximate Markov chain, where progressively removing noise in \(z_T\) through the reverse process, ultimately restoring the noise-free latent code \(z_0\) after \(T\) iterative steps:
\begin{gather}
    \label{eq:reverse_ddpm2}
     \mathbf{z}_{t-1} = \frac{1}{\sqrt{1-\beta_t}}(\mathbf{z}_t - \frac{\beta_t}{\sqrt{1-\alpha_t}}\boldsymbol{\epsilon}_\theta(\mathbf{z}_t, t)) + \sigma_t\mathbf{\epsilon},
\end{gather}
where \(\boldsymbol{\epsilon}_\theta\) is a time-conditioned U-Net \cite{ronnebergerUnetConvolutionalNetworks2015}, tasked with predicting the noise component in \(z_t\) at each step \(t\). The parameters \(\theta\) within \(\boldsymbol{\epsilon}_\theta\) are fine-tuned by minimizing a noise prediction loss:
\begin{equation}
L(\theta)=\mathbb{E}_{t,z_0, \epsilon}\left[\left\|\boldsymbol{\epsilon}-\boldsymbol{\epsilon}_\theta\left(\boldsymbol{z}_t,t\right)\right\|^2\right],
\end{equation}
where \(\boldsymbol{\epsilon}\) denotes the noise introduced during the forward process as described in Eq. \ref{eq:forward_ddpm}.

\subsection{Latent Consistency Models}

Latent Consistency Models (LCMs)~\cite{luoLatentConsistencyModels2023}, are a specialized form of Consistency Models (CMs)~\cite{songConsistencyModels2023} that significantly accelerate the generation speed of LDMs. In LCMs, the consistency function $f(z_t, t)$ ensures that each anchor point \(z_t\) in the sampling trajectory can be accurately mapped back to the initial latent code \(z_0\), thereby ensuring self-consistency within the model. The consistency function is defined as follows:

\begin{align}
    f(x, t) = c_\text{skip}(t) x + c_\text{out}(t) F(x, t)\label{eq:consistencyfunc},
\end{align}
where $c_\text{skip}(t)$ and $c_\text{out}(t)$ are differentiable functions designed to ensure the differentiability of \(f(x, t)\) with conditions \(c_\text{skip}(0) = 1\) and \(c_\text{out}(0) = 0\). The efficacy of $f(x, t)$ is measured through the following optimization objective:

\begin{equation}
\label{eq:consistency_loss}
\min_{\theta,\theta^{-};\phi} \mathbb{E}_{z_0,t} \left[d\left(f_{\theta}(z_{t+1}, t+1), f_{\theta^{-}}(\hat{z}^{\phi}_{t}, t)\right)\right],
\end{equation}
where \(f_{\theta}\) denotes a consistency function parameterized by a trainable neural network, and \(f_{\theta^-}\) is updated at a slow decay rate \(u\) to adjust parameters within \(f_{\theta}\). The variable \(\hat{z}^{\phi}_{t}\) represents a one-step estimate of \(z_t\) obtained through the sampler \(\phi\) from \(z_{t+1}\).

\section{Method}

In this section, we introduce \textbf{ZePo}, a zero-shot framework for portrait stylization that operates within four sampling steps. Our framework leverages Latent Consistency Models (LCMs), a variant of Stable Diffusion that distilled with the consistent objective (Eq. \ref{eq:consistency_loss}). ZePo capitalizes on the observation that LCMs not only significantly reduce the number of sampling time steps required for generating images but also efficiently extract representative features from noisy images, which we term Consistency Features. Utilizing the Consistency Features extracted from both source and reference images, we seamlessly integrate these features into the image generation process through our proposed Style Enhancement Attention Control module. This integration allows for subtle yet effective stylization adjustments. Ultimately, with just four sampling steps, our framework is capable of synthesizing high-quality stylized portraits that faithfully capture the style of the reference image. The overall architecture of our framework is depicted in Figure \ref{fig:framework}.

\subsection{Consistency Features}

\begin{figure}[!h]
        \centering
        \includegraphics[width=1\linewidth]{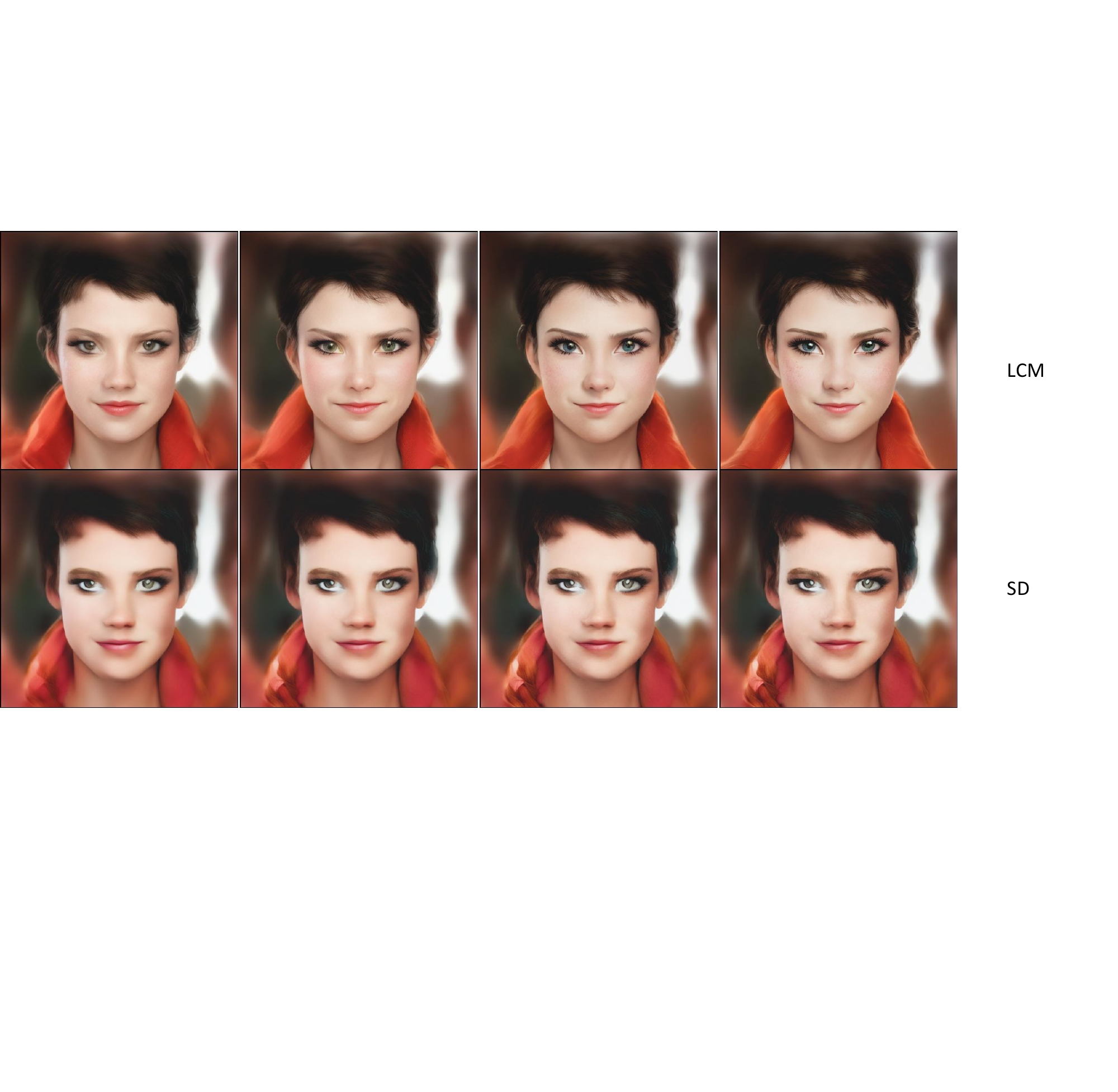}
        \caption{The results of one-step denoising with different noise levels (time-step), different noise addition methods (DDIM Inversion and Forward Process), and different models (SD and LCM) are examined. (a) DDIM Inversion + SD. (b) Forward Process + SD. (c) Forward Process + LCM.  } 
        \label{fig:one_step_denoise}
\end{figure}

The primary purpose of employing DDIM Inversion \cite{songDenoisingDiffusionImplicit2020} is to derive a series of anchor points \(\{z_t\}\) that facilitate the reconstruction of the original image \(z_0\), where each anchor \(z_t\) is capable of recovering \(z_0\) with better accuracy. As illustrated in Fig. \ref{fig:one_step_denoise} (a) (b), utilizing the noisy latent \(z_t\) derived from the forward process in Eq. \ref{eq:forward_ddpm}, tends to yield a predicted \(z_0\) that is blurry and lacks high-frequency details. In contrast, the noisy latent $z_t$ post DDIM Inversion can estimate $z_0$ with enhanced accuracy.
The optimization objective (Eq. \ref{eq:consistency_loss}) of LCMs is aims to minimize the disparity between the outputs of consistency function in adjacent samples, which corresponds to the distance between one-step predictions of the model for \( z_0 \). It is observed that this objective endows LCMs with superior one-step predictive capabilities for \( z_0 \). As depicted in Figure \ref{fig:one_step_denoise} (c), the noise level during forward process is relatively low, particularly for time steps \( t \leq 300 \), the estimated \( \hat{z_0} \) by LCM exhibits clearer and more consistent details compared to the original \( z_0 \). This suggests that LCMs can effectively extract representative features from a noisy image, which is referred to Consistency Features.
Inspired by this capability, we propose leveraging the Consistency Features extracted from both source and reference images for portrait stylization. This method effectively replaces the time-consuming DDIM Inversion process, offering a more efficient pathway to achieving high-quality portraits stylization.

Given a source image \(I^{src}\) and a reference image \(I^{ref}\), initially, a pre-trained VAE encoder \(\mathcal{E}\) encodes them into latent codes \(z^{src}\) and \(z^{ref}\), respectively. Subsequently, a forward process (Eq. \ref{eq:forward_ddpm}) is applied to introduce noise to these latent codes in a single step, defined as follows:
\begin{equation}
\begin{aligned}
z^{src}_t &= \sqrt{{\alpha}_t} z^{src}_0 + \sqrt{1-{\alpha}_t} \epsilon \\
z^{ref}_t &= \sqrt{{\alpha}_t} z^{ref}_0 + \sqrt{1-{\alpha}_t} \epsilon,
\end{aligned}
\end{equation}

where \(t\) represents a smaller time-step, and \(\epsilon \sim \mathcal{N}\left(0,\boldsymbol{I}\right)\).
Finally, the noisy latent codes \(z^{src}_t\) and \(z^{ref}_t\) are inputted into the noise prediction network \(\epsilon_{\theta}\) of the LCMs, from which the consistency features \(\{f_l^{src}\}\) and \(\{f_l^{ref}\}\) of the source and reference images at each transformer layer $l$ of  \(\epsilon_{\theta}\) are extracted. This process is formalized as:

\begin{equation}
(\{f_l^{src}\}, \{f_l^{ref}\}) = \epsilon_{\theta}((z^{src}_t, z^{ref}_t),t,c),
\end{equation}

where \(c\) denotes the textual condition.

In contrast to previous approaches \cite{tumanyanPlugandPlayDiffusionFeatures2023,caoMasaCtrlTuningFreeMutual2023} which necessitate feature injection to align with the current generation time step, our proposed consistency features exhibit flexibility in this regard. They are not bound by the requirement to match the current generation time step. Thus, the extracted consistency features can seamlessly integrate into the generation process at any time step, ensuring their consistent contribution throughout various stages of the generation process. We demonstrate the impact of feature extraction at different time steps on the generated results in Figure \ref{fig:ablation_fixstep}.

\subsection{Style Enhancement Attention Control}

\noindent
\textbf{Attention Control.} Attention Control (AC) replaces the key and value features in the target image generation branch with those derived from the source image reconstruction branch. Leveraging the self-attention mechanism, AC adaptively aggregates features from the reference image, thereby preserving both semantic and structural information from the source image. However, the incorporation of AC significantly impacts the speed of image generation in existing methods. We conducted a comparative analysis, measuring the time required for image generation with and without AC under identical step settings. As presented in Table \ref{tab:ac_time}, which indicates approximately a 30\% increase in time consumption when AC is employed.
\begin{table}[!h]
\caption{The image generation speeds with and without Attention Control (AC) at different time steps.}
\begin{tabular}{llll}
\hline
       & T=50      & T=25      & T=10      \\ \hline
w/o AC & 06.55     & 03.00     & 01.22     \\
W AC   & 08.61     & 04.17     & 01.64     \\ \hline
\end{tabular}
\label{tab:ac_time}
\end{table}

\noindent
\textbf{Feature Merge.} 
In Vision Transformers, there exists redundancy in tokens, and pruning these redundant tokens during inference can lead to a model with faster inference speed \cite{bolyaTokenMergingYour2023}. Similar techniques have been investigated within the diffusion model framework, which extensively employs self-attention modules. Merging redundant features in diffusion models has been shown to significantly enhance the speed of image generation without compromising the quality of the generated images \cite{bolyaTokenMergingFast2023}. Building upon this observation, we propose leveraging the token merge technique to merge redundant feature sequences before to attention control, thereby reducing the length of features from \(N\) to \(N/2\) or less, in which we randomly sample one feature from each 2x2 patch as the target and merge the \(50\%\) most similar features from the source into the target, resulting in a final set of features where only half of the original features remain.
In contrast to the approach outlined in \cite{bolyaTokenMergingFast2023}, which necessitates the un-merging of merged token sequences to restore the original length of token sequences, our method exclusively merges the consistency features inputted into attention control. This targeted merging strategy helps circumvent errors that may arise during the un-merging process.

\noindent
\textbf{Style Enhancement Attention Control}
We denote the merged consistency features at layer \(l\) as \((\hat{f^{src}_l}, \hat{f^{ref}_l})\). Upon entering the Attention Control mechanism, these merged features are individually mapped to the key \((K^{src}, K^{ref})\) and value \((V^{src}, V^{ref})\) features within the self-attention module.
In contrast to the conventional AC methods that directly replace key and value features, we introduce a Style Enhancement Attention Control (SEAC) mechanism. Specifically, we concatenate \( (K^{src}, K^{ref}) \) and \( (V^{src}, V^{ref}) \) from the source and reference images into a unified set of key and value features. Moreover, we enhance \( K^{ref} \) by multiplying it with a Style Enhancement coefficient \( \lambda \), yielding a new set of key and value features as follows:
\begin{equation}
\begin{aligned}
    K^{sr} &= \text{Concat}(K^{src}, \lambda \cdot K^{ref}) \in \mathbb{R}^{B,N,D}, \\
    V^{sr} &= \text{Concat}(V^{src}, V^{ref}) \in \mathbb{R}^{B,N,D}.
\end{aligned}
\end{equation}

Subsequently, the key feature \( K^{sr} \) and the query feature \( Q^{tgt} \in \mathbb{R}^{B,N,D} \) from the target image are utilized to compute a self-attention map \( A \) given by:

\begin{equation}
A = \text{SoftMax}\left(\frac{Q^{tgt} \cdot {K^{sr}}^T}{\sqrt{d}}\right) \in \mathbb{R}^{B,N,N},
\end{equation}
where \( d \) represents the dimensionality of the query and key features. Finally, the self-attention map \( A \) is applied to the value feature \( V^{sr} \) to derive the final output \( \phi^{tgt} \) as follows:
\begin{equation}
\phi^{tgt} = A \cdot V^{sr}.
\end{equation}
Hence, SEAC can effectively assess the similarity between the query features \( Q^{tgt} \) and the combined key features \( (K^{src}, K^{ref}) \), enabling the adaptive aggregation of value features from \( (V^{src}, V^{ref}) \). Additionally, the lengths of the query, key, and value features utilized in the attention computation are all \( N \), ensuring consistency in the computational cost of attention control compared to the original self-attention mechanism. The comprehensive pipeline of the Style Enhancement Attention Control is illustrated in Fig. \ref{fig:framework} (right).

Building on the consistency feature and Style Enhancement Attention Control, we introduce the rapid portrait stylization framework \textbf{ZePo}. This framework synthesizes stylized faces in four sampling steps, detailed in Algorithm \ref{alg1}.

\begin{algorithm}[htb]  
  \caption{Zero-shot Portrait Stylization}  
  \label{alg1}  
  \begin{algorithmic}[1]  
    \REQUIRE  \hskip\algorithmicindent \\
    \hskip\algorithmicindent Distillated Diffusion Model $\epsilon_\theta$,  Encoder $\mathcal{E}$, Decoder $\mathcal{D}$;  \\
    \hskip\algorithmicindent Prompt condition $c$, Guidance scale $s$, Sample steps $T$ ; \\
    \hskip\algorithmicindent Reference image $x^{ref}$, Source image $x^{src}$, Consistency feature step $\tau$,;  \\
		\STATE $z^{ref}_0,z^{src}_0 \longleftarrow \mathcal{E}(x^{ref},x^{src})$;
		\STATE Sample noise $\epsilon \longleftarrow \mathcal{N}\left(0, \mathbf{I}\right)$;
  		\STATE $(z^{ref}_\tau,z^{src}_\tau) \longleftarrow$ Forward$((z^{ref}_0,z^{src}_0),\tau,\epsilon)$;
	    \STATE $(\{f^{ref}_l\},\{f^{src}_l\}) \leftarrow \epsilon_{\theta}\left( (z^{ref}_\tau,z^{src}_\tau),\tau,c,s \right)$;
        \STATE ${z}^{tgt}_0\leftarrow {z}^{src}_0$
        \STATE $t=T$
        \REPEAT
            \STATE $t = t-1$
            \STATE Sample noise $\epsilon \longleftarrow \mathcal{N}\left(0, \mathbf{I}\right)$;
            \STATE $z^{tgt}_t \longleftarrow$ Forward$(z^{src}_0,t,\epsilon)$;
            \STATE $\epsilon^{tgt} \leftarrow  \epsilon_{\theta}\left( z^{tgt}_t,t,c,s,(\{f^{ref}_l\},\{f^{src}_l\}) \right)$ ;
            \STATE ${z}^{tgt}_0\leftarrow$ Prediction$(z^{tgt}_t,t,\epsilon^{tgt})$;            
		\UNTIL {$t < 0$} 
        \RETURN $x^{tgt} \longleftarrow \mathcal{D}({z}^{tgt}_0)$
  \end{algorithmic}  
\end{algorithm}

\begin{figure*}[!t]
	\centering
	\subfloat[Style]{\includegraphics[width=.12\linewidth]{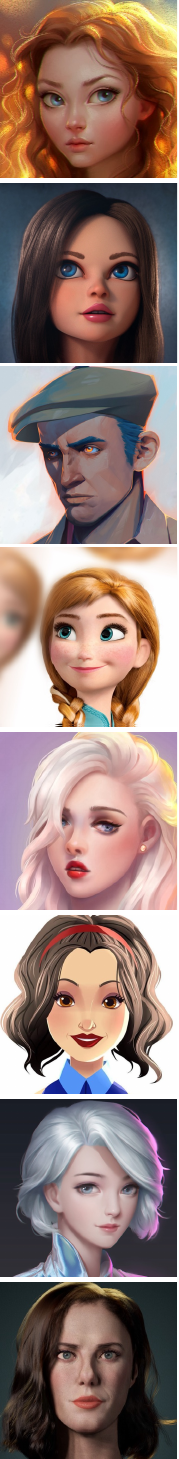}}
	\subfloat[Content]{\includegraphics[width=.12\linewidth]{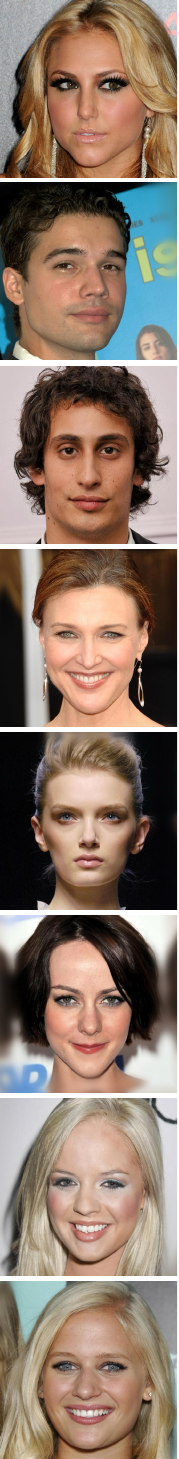}}
	\subfloat[Ours]{\includegraphics[width=.12\linewidth]{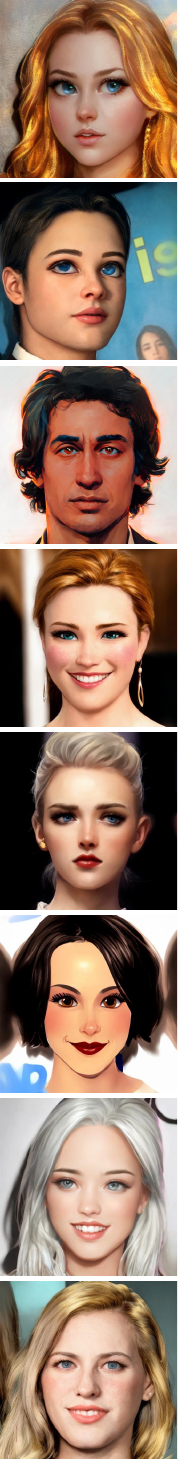}}
	\subfloat[VCT \cite{chengGeneralImagetoImageTranslation2023}]{\includegraphics[width=.12\linewidth]{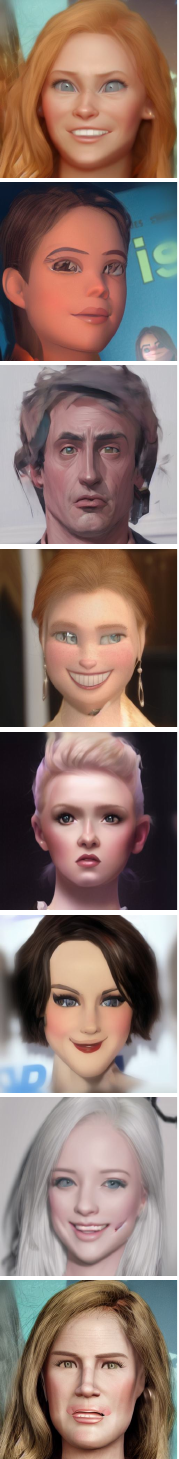}}
        \subfloat[InST \cite{zhangInversionbasedStyleTransfer2023}]{\includegraphics[width=.12\linewidth]{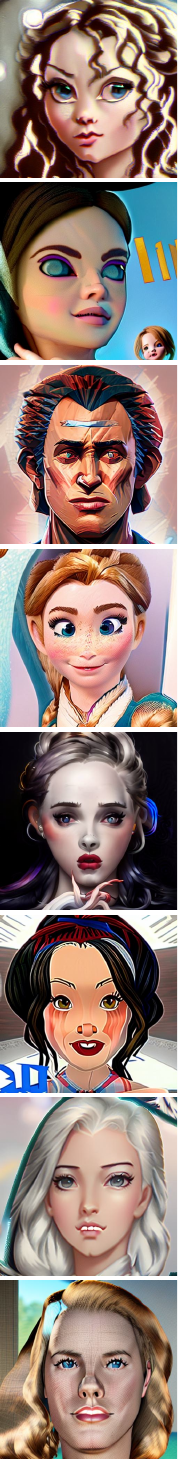}}
        \subfloat[NADA \cite{galStyleGANNADACLIPguidedDomain2022}]{\includegraphics[width=.12\linewidth]{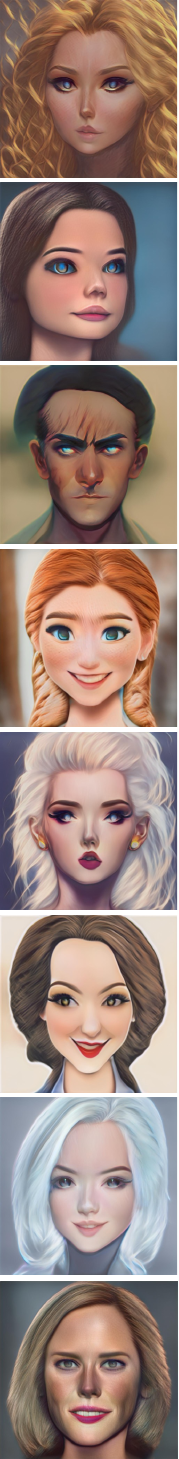}}
        \subfloat[DynaGAN \cite{kimDynaGANDynamicFewshot2022}]{\includegraphics[width=.12\linewidth]{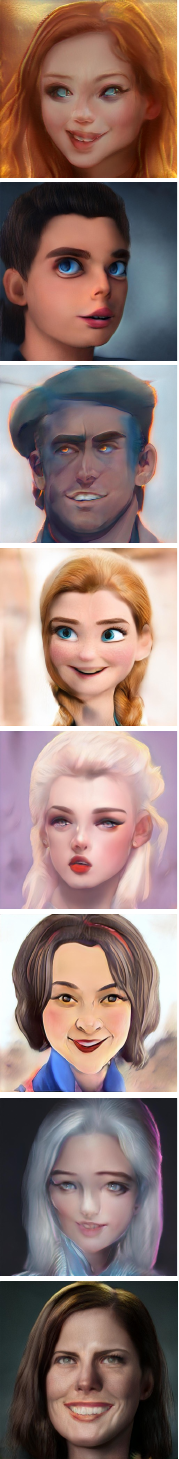}}
        \subfloat[JoJoGAN \cite{chongJojoganOneShot2022}] {\includegraphics[width=.12\linewidth]{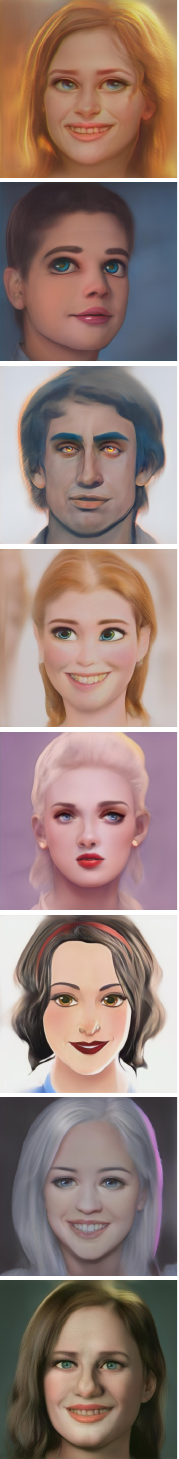}}
        
	\caption{Qualitative comparisons with conventional portrait stylization baselines. (a) and (b) are the input reference image and content image, respectively, while (c-h) are the stylized portraits synthesized by different baselines.}
        \label{fig:comp}
\end{figure*}

\section{Experiments}

\noindent
\textbf{Implementation Details.} Our experiments use Latent Consistency Models (LCMs), a variant of acceleration-distilled Stable Diffusion, with the LCM sampler. Each stylized image is synthesized in four sampling steps, using “head” as the conditional text prompt and a classifier-free guidance scale of $2$. The style enhancement coefficient $\lambda$ is $1.2$. Experiments are run on a single NVIDIA 4090 GPU. Reference images are from the AFHQ dataset \cite{liuBlendganImplicitlyGan2021}, and content images are from the CelebA-HQ dataset \cite{karrasProgressiveGrowingGANs2018}, all processed at 512 × 512 resolution.

\begin{table*}[hbtp]
\caption{Quantitative comparison with conventional portrait stylization baselines. The best and second best of each metrics will be highlighted in \textbf{boldface} and \underline{underline} format, respectively. }

\label{tab:comp}

\begin{tabular}{ccccccc}
\hline
                                 & Methods        & LPIPS $\downarrow$   & CLIP-IQA $\uparrow$ & Style $\downarrow$   & Fine-tuning(s) $\downarrow$ & Inference(s) $\downarrow$ \\ \hline
\multirow{3}{*}{StyleGAN-Based}  & JoJoGAN \cite{chongJojoganOneShot2022}       & 0.550 & 0.538  & \underline{3.742}  & \underline{48.524}         & \underline{0.052}       \\
                                 & DynaGAN \cite{kimDynaGANDynamicFewshot2022}      & 0.588 & 0.555  & \textbf{2.810}  & 1156.822       & \textbf{0.041}        \\
                                 & NADA \cite{galStyleGANNADACLIPguidedDomain2022} & 0.561 & 0.566  & 4.813  & 155.321        & 0.091       \\ \hline
\multirow{3}{*}{Diffusion Based} & InST \cite{zhangInversionbasedStyleTransfer2023}         & 0.564 & \underline{0.727}  & 5.775 & 2007.966       & 6.932       \\
                                 & VCT \cite{chengGeneralImagetoImageTranslation2023}         & \underline{0.348}  & 0.467  & 5.887 & 374.117        & 37.850      \\
                                 & \textbf{Our}           & \textbf{0.261}   & \textbf{0.858}  & {5.213} & \textbf{0}             & {0.684}       \\ \hline
\end{tabular}

\end{table*}

\subsection{Qualitative Comparison}

\noindent
\textbf{Baselines.}  
To evaluate our method’s efficacy, we conducted extensive comparative experiments against current state-of-the-art (SOTA) few-shot adaptation techniques. We included StyleGAN-based approaches such as JoJoGAN \cite{chongJojoganOneShot2022}, StyleGAN NADA (NADA) \cite{galStyleGANNADACLIPguidedDomain2022}, and DynaGAN \cite{kimDynaGANDynamicFewshot2022}. Additionally, we compared our method with diffusion-based methods like InST \cite{zhangInversionbasedStyleTransfer2023} and VCT \cite{chengGeneralImagetoImageTranslation2023}. All stylized outputs were generated using the open-source implementations provided by the authors. Figure \ref{fig:comp} qualitatively compares various methods. (a) presents reference artistic portraits, while (b) displays the original natural faces. And (c) illustrates the results of our ZePo, and the subsequent columns showcase outputs from various competing models. In Figure \ref{fig:comp} (f-h), StyleGAN-based methods effectively transfer style but often result in over-stylization, altering facial poses as seen in the fourth and fifth rows for NADA \cite{galStyleGANNADACLIPguidedDomain2022}. JoJoGAN \cite{chongJojoganOneShot2022} excels in stylization but struggles with content consistency, especially in preserving backgrounds. Among diffusion-based methods, InST \cite{zhangInversionbasedStyleTransfer2023} tends to overfit, while VCT \cite{chengGeneralImagetoImageTranslation2023} better balances style transformation and content retention but changes expressions. Our method excels in preserving local details like facial features and hair texture, maintaining consistent facial characteristics, exemplified by the preservation of earrings in the first and fourth rows.

\subsection{Quantitative Comparison.} 
To demonstrate the superior quality and efficiency of our method in portrait art synthesis, we conducted quantitative comparisons with existing SOTA methods.

\noindent
\textbf{Metric.} To objectively assess the effectiveness of our proposed method, we employed LPIPS \cite{zhangUnreasonableEffectivenessDeep2018} for content preservation and VGG Style loss \cite{gatysImageStyleTransfer2016a} for stylization evaluation. We observed that style loss predominantly focuses on external texture styles, which does not effectively capture the intrinsic style of images. Consequently, we propose the adoption of the non-referential evaluation metric, CLIP-IQA \cite{wangExploringClipAssessing2023}, for a more comprehensive assessment of image quality. CLIP-IQA leverages the CLIP model \cite{radfordLearningTransferableVisual2021}, pre-trained on a large-scale text-image paired dataset, as an image feature extractor. Then, this method evaluates the overall image quality through different text prompts that relate to image quality and aesthetics.

\noindent
\textbf{Evaluation.} For quantitative assessment, we randomly selected 10 style images and 10 content images, generating a total of 100 stylized images for each baseline. The quantitative results are presented in Table \ref{tab:comp}. Our method outperformed other techniques, achieving the best scores on both LPIPS and CLIP-IQA metrics. A lower LPIPS score indicates superior content preservation by our method, while a higher CLIP-IQA score reflects our method's ability to synthesize images with better overall quality and visual appeal. Additionally, our style score was the highest among methods based on diffusion models.
We also evaluated the fine-tuning and inference times required by each method, as shown in Table \ref{tab:comp}. Previous methods demand extended fine-tuning periods, and diffusion-based methods exhibit longer inference times. For example, InST \cite{zhangInversionbasedStyleTransfer2023} requires around 7 seconds to synthesize one stylized image, while VCT \cite{chengGeneralImagetoImageTranslation2023} needs 37 seconds due to Null-text text inversion \cite{mokadyNULLTextInversionEditing2023a}. Our framework, utilizing a zero-shot approach, eliminates the need for additional fine-tuning. By incorporating Style Enhancement Attention Control, we have reduced the inference time to approximately 0.6 seconds, enhancing the practicality of our method.

\begin{figure}[!h]
        \centering
        \includegraphics[width=1\linewidth]{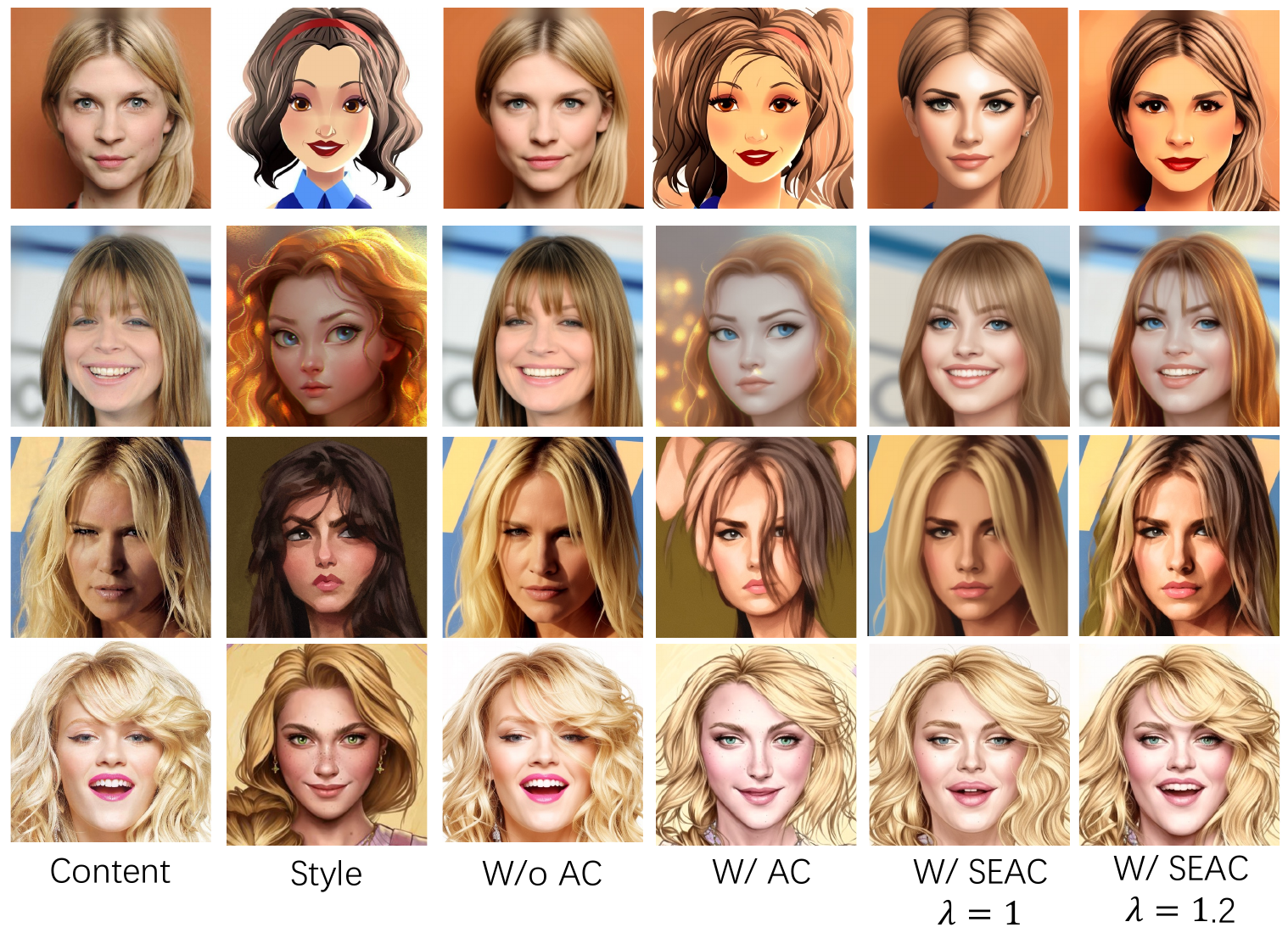}
        \caption{Ablation experiments on Attention Control. }
        \label{fig:ablation_ac}
\end{figure}

\subsection{Ablation Study}

\noindent
\textbf{Attention Control.} We conducted extensive ablation experiments to verify the effectiveness of the proposed Style Enhancement Attention Control (SEAC). Figure \ref{fig:ablation_ac} presents the ablation results using different Attention Control (AC) methods. 
Excluding AC results in merely the reconstruction of content images, lacking any substantive stylization. Conversely, the use of conventional AC often leads to over-stylization and the loss of critical content details. In contrast, our proposed Style Enhancement Attention Control (SEAC) maintains the integrity of content information while imparting a more subtle stylization effect. Additionally, the Style Enhancement (SE) coefficient effectively controls the strength of stylization. By adjusting the SE coefficient to 1.2, the stylization effect is notably enhanced, thus affirming the capability of SEAC to maintain a balance between content preservation and the desired level of stylization.

\begin{figure}[!h]
        \centering
        \includegraphics[width=1\linewidth]{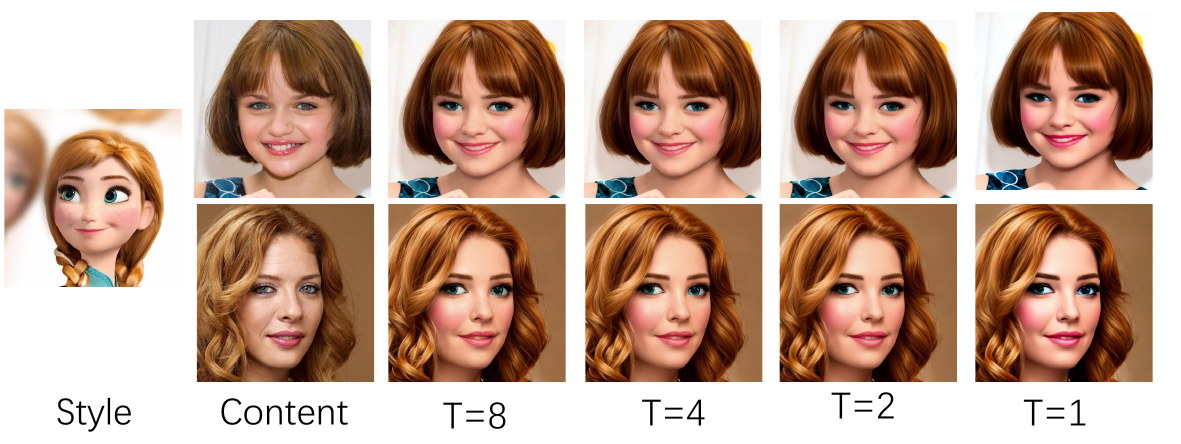}
        \caption{Ablation experiment on sampling time steps. Our method produces satisfactory stylized results with just one sampling step. Increasing the number of sampling steps further enhances the detail in the synthesized results.}
        \label{fig:ablation_step}
\end{figure}

\noindent
\textbf{Inference Steps} Figure \ref{fig:ablation_step} illustrates the results produced by our method at various sampling steps. Notably, our method can generate satisfactory stylized outcomes with just a single sampling step, and further increasing the number of sampling steps refines the detail of the synthesized images. As indicated in Table \ref{tab:ablation_step}, enhancing the number of sampling steps leads to higher CLIP-IQA scores. However, this increment also results in a slight decline in content preservation and inference speed. To strike an optimal balance among stylization quality, content preservation, and inference efficiency, we established the sampling steps at four for all experiments.
Additionally, we validated the effectiveness of the Feature Merge (FM) technique. As depicted in Table \ref{tab:ablation_step}, implementing FM reduces the time required to synthesize images by 20\%, without significantly compromising the quality of the generated images. This demonstrates that the feature merge technique not only enhances efficiency but also maintains high-quality stylization outcomes.

\begin{table}[]
\caption{Ablation experiment on different sampling time steps T and  Feature Merge (FM). Increasing the sampling time steps can improve the CLIP-IQA score and reduce style loss. Employing FM can enhance the model's inference speed without significantly affecting the quality of the images.} 
\begin{tabular}{cccccc}
\hline
                                      &   Step   & CLIP-IQA $\uparrow$ & LPIPS $\downarrow$& Style $\downarrow$& Inference(s) $\downarrow$\\ \hline
\multirow{4}{*}{\begin{tabular}[c]{@{}l@{}}W/ \\ FM\end{tabular}}     & T=25 & 0.797  & 0.470 & \textbf{1.628} & 2.862        \\
                                      & T=8  & 0.835  & 0.409 & 1.848 & 1.049        \\
                                      & T=4  & 0.823  & 0.376 & 1.993 & 0.634        \\
                                      & T=2  & 0.781  & 0.326 & 2.181 & 0.416        \\
                                      & T=1  & 0.753  & 0.242 & 2.357 & \textbf{0.304}        \\
                                      \hline
\multirow{4}{*}{\begin{tabular}[c]{@{}l@{}}W/o \\ FM\end{tabular}} & T=25 & 0.778  & 0.440 & 1.747 & 3.434        \\
                                      & T=8  & \textbf{0.844}  & 0.413 & 1.994 & 1.232        \\
                                      & T=4  & 0.817  & 0.368 & 2.094 & 0.724        \\
                                      & T=2  & 0.779  & 0.313 & 2.228 & 0.467        \\ 
                                      & T=1  & 0.766  & \textbf{0.226} & 2.411 & 0.355        \\
                                      \hline
\end{tabular}
        \label{tab:ablation_step}
        
\end{table}

\begin{figure}[!h]
        \centering
        \includegraphics[width=1\linewidth]{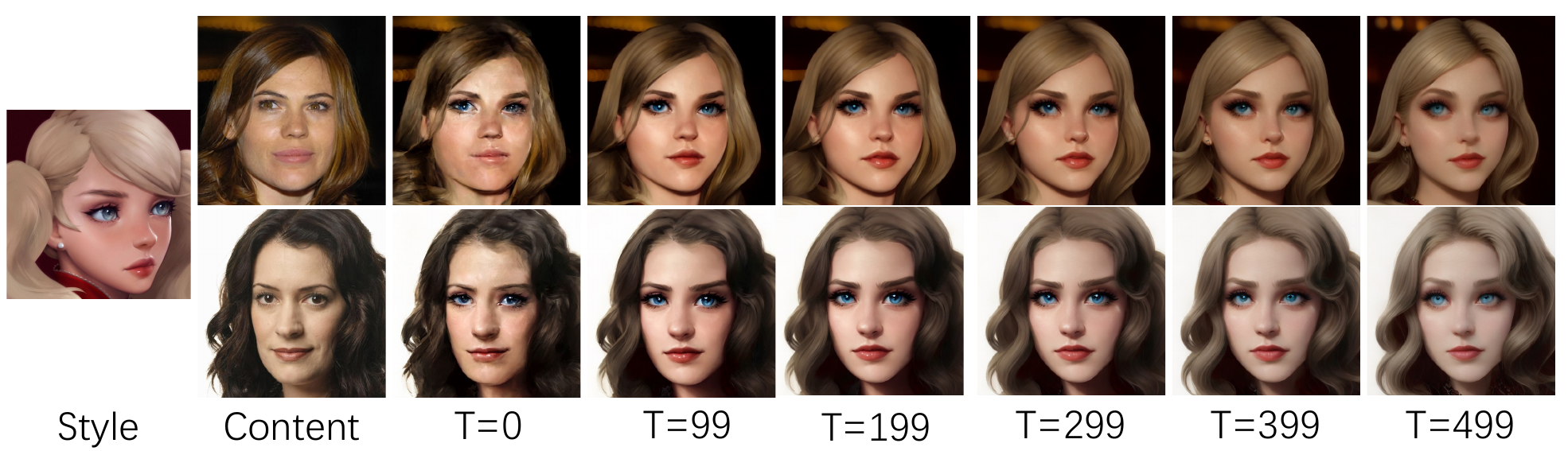}
        \caption{Ablation experiment on the time-step $T$ during the forward noise addition process in consistency feature extraction. } \label{fig:ablation_fixstep}
\end{figure}

\noindent
\textbf{Consistency Features.} We performed an ablation study on the use of a fixed time step for consistent feature extraction in Figure \ref{fig:ablation_fixstep}. Contrary to initial expectations, extracting features directly from the input image without the addition of noise results in the inability of the model to discern content and style features effectively. This phenomenon is consistent with the behavior of the consistency equation (Eq. \ref{eq:consistencyfunc}) at \( t=0 \), where it merely outputs \( z_0 \) without any processing through the network. Consequently, during the consistency model distillation process, the noise prediction network \( \epsilon_\theta \) has not learned to process inputs at \( t=0 \) and therefore fails to extract features directly from \( z_0 \). Based on these ablation insights, we have set the fixed time step for consistency feature extraction to $99$, enabling the extraction of more distinct consistency features.

\begin{figure}[!ht]
	\centering
	\includegraphics[width=.92\linewidth]{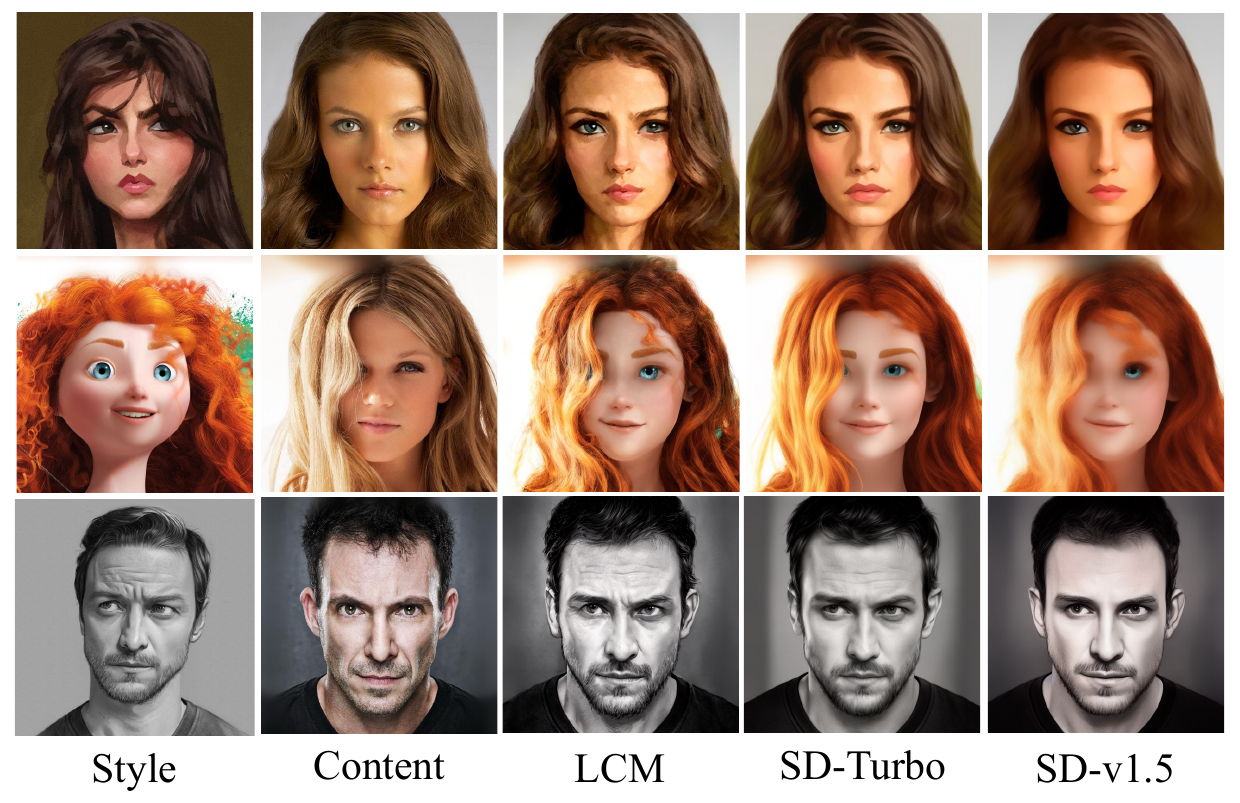}
	\caption{Portrait stylization results with different models.}
        \label{fig:model_comp}
\end{figure}

\noindent
\textbf{Base Models.} To demonstrate our method’s effectiveness, we present stylization results using various base models in Figure \ref{fig:model_comp}, including Latent Consistency Models (LCMs) \cite{luoLatentConsistencyModels2023}, SD-Turbo \cite{sauer2023adversarial}, and SD-v1.5 \cite{rombachHighresolutionImageSynthesis2022}. SD-Turbo is an accelerated version of SD-v2.1 \cite{rombachHighresolutionImageSynthesis2022} through distillation. The results show that the undistilled SD-v1.5 model produces blurry images with a lack of high-frequency details, consistent with our observations in Figure 3. In contrast, the distilled SD-Turbo extracts more representative features, resulting in better stylization. Both LCMs and SD-Turbo generate excellent stylized results, but LCMs, which use consistency loss for distillation, capture high-frequency details better than SD-Turbo, which uses Adversarial Diffusion Distillation \cite{sauer2023adversarial}. This highlights the effectiveness of LCMs in extracting consistent features.

\section{Conclusion}

In this study, we introduce ZePo, a framework that rapidly generates stylized portraits without needing fine-tuning on specific samples or DDIM Inversion for input images. This allows for high-quality stylization within four sampling steps, reducing inference time to 0.6 seconds. ZePo utilizes a Consistency Features extraction strategy, leveraging a pre-trained diffusion model to extract multi-scale Consistency Features from content and reference images. Our style Enhancement Attention Control module adaptively fuses these features, enabling adjustable stylization intensity through the style enhancement coefficient. Additionally, our feature merge technique reduces redundant consistency features, significantly lowering the computational cost of attention control and enhancing sampling speed. Extensive experiments show that ZePo synthesizes high-quality stylized results while preserving the content integrity of the source image, surpassing the performance of existing advanced methods.

\newpage
\newpage
\begin{acks}
This work is partially funded by National Natural Science Foundation of China (Grant No U21B2045, U20A20223, 32341009, 62206277), Beijing Nova Program (20230484276), and Youth Innovation Promotion Association CAS (Grant No. 2022132).

\end{acks}

\bibliographystyle{ACM-Reference-Format}
\bibliography{zepo}

\end{document}